\documentclass[runningheads]{llncs}
\usepackage{comment}
\usepackage{graphicx}
\usepackage{amsmath}
\usepackage{cite}
\usepackage{enumitem}

\usepackage{caption}
\usepackage{subcaption}
\usepackage{algorithmic}
\usepackage[noend,ruled,noline,linesnumbered]{algorithm2e}
\graphicspath{{figures/}}
\usepackage{verbatim}
\usepackage[utf8]{inputenc}
\usepackage{amsmath,amssymb,amsfonts}
\usepackage{algorithmic}
\usepackage{graphicx}
\usepackage{graphics}
\usepackage{lipsum}
\usepackage{textcomp}
\usepackage{xcolor}
\usepackage{hyperref}
\hypersetup{
    colorlinks=true,
    linkcolor=blue,
    filecolor=magenta,      
    urlcolor=cyan,
}
\usepackage{hypcap}
\usepackage{float}
\usepackage{epstopdf}
\renewcommand{\thetable}{\arabic{table}}

\begin{document}
\title{Automatic Signboard Detection and Localization in Densely Populated Developing Cities}
\titlerunning{Automatic Signboard Localization}
%
\author{Md. Sadrul Islam Toaha \and
Sakib Bin Asad \and
Chowdhury Rafeed Rahman \and
S.M. Shahriar Haque \and
Mahfuz Ara Proma \and
Md. Ahsan Habib Shuvo \and
Tashin Ahmed \and
Md. Amimul Basher
}
\authorrunning{Toaha et al.}
%
\institute{United International University, Dhaka, Bangladesh}
\maketitle              
\begin{abstract}
Most city establishments of developing cities are digitally unlabeled because of the lack of automatic annotation systems. Hence location and trajectory services such as Google Maps, Uber etc remain underutilized in such cities. Accurate signboard detection in natural scene images is the foremost task for error-free information retrieval from such city streets. Yet, developing accurate signboard localization system is still an unresolved challenge because of its diverse appearances that include textual images and perplexing backgrounds. We present a novel object detection approach that can detect signboards automatically and is suitable for such cities. We use Faster R-CNN based localization by incorporating two specialized pretraining methods and a run time efficient hyperparameter value selection algorithm. We have taken an incremental approach in reaching our final proposed method through detailed evaluation and comparison with baselines using our constructed SVSO (Street View Signboard Objects) signboard dataset containing signboard natural scene images of six developing countries. We demonstrate state-of-the-art performance of our proposed method on both SVSO dataset and Open Image Dataset. Our proposed method can detect signboards accurately (even if the images contain multiple signboards with diverse shapes and colours in a noisy background) achieving 0.90 $mAP$ (mean average precision) score on SVSO independent test set. Our implementation is available at: \textit{https://github.com/sadrultoaha/Signboard-Detection}

\end{abstract}

\section{Introduction}
\begin{figure}[h]
    \centering
    \includegraphics[width=0.8\textwidth]{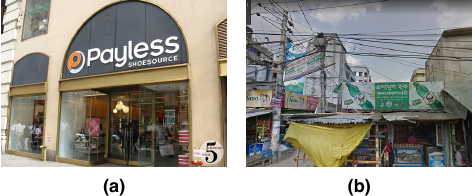}
    \caption{Problem complexity scenario: \textbf{(a)}: Signboard characteristics of a developed city. \textbf{(b)}: Scenario of developing or less developed city which reveals the complexity of our case.}
    \label{fig:dataprob}
\end{figure}

Automatic signboard detection task has obtained its importance in computer vision and information retrieval field with many practical and relevant applications. Some of them are - store classification, automatic annotation of establishments in web map, information extraction from signboards and unlawful signboard monitoring \cite{storeclassification, socr, building}. Nowadays, location based web services such as Google Maps, UBER and Foodpanda are growing fast. Currently, these companies require manual labeling of street establishments such as shops, hospitals, banks and other commercial landmarks \cite{ourmap, ontology}. Thus, the establishment annotation process is still inefficient and in some cases inaccurate \cite{largewebsite}. In this case, an automatic establishment annotation system can solve this problem, as signboards are vital information source for street establishments. The automatic establishment annotation problem can be divided into the following three sub tasks: signboard detection, information extraction and web map annotation. However, extracting information from signboard image is still an open problem due to large variation of signboards from one country to another. Variations include noise, brightness, captured angle and background type \cite{ socr, building, koreantextoutdoor, robust, realtime, largescale, imagevideo}. Therefore, it is important to localize the signboards in an input image for the development of an efficient establishment annotation system. Existing researches on signboard localization \cite{storeclassification, building, robust, generic, signboardDANN} show limitations in their capability when faced with challenging urban features of developing countries \cite{iop} such as - (i) diverse shapes and appearances of signboards, (ii) presence of pillars, trees and wires across the signboards, (iii) identical manifestation of signboards, (iv) deformed and poor quality signboards and so on. In \textbf{Figure \ref{fig:dataprob}a}, the image depicts the signboard characteristics of a developed city, while \textbf{Figure \ref{fig:dataprob}b} shows the scenario of developing or less developed cities.
The foremost goal of this research is to solve the first step of the automatic establishment annotation problem by detecting and localizing signboards from an image in the context of densely populated developing cities.

 In relevant researches, authors performed signboard detection in three types of approaches: (i) text segmentation method (TSM), (ii) classical image processing and edge detection (CIPED) and (iii) segmentation based learning algorithm (SBLA). However, existing works commonly focus on one or two of the mentioned approaches. Applying \textbf{TSM} without detecting the signboard is a common strategy for retrieving  signboard information \cite{socr, koreantextoutdoor} aiming to minimize the system complexity. Authors in \cite{socr} used Maximally Stable Extremely Region (MSER) algorithm to detect signboard region by segmenting texts for information extraction. They showed that their multi scale approach combined with TSM is able to recognize signboard texts successfully. A system according to TSM technique for signboard text recognition and translation was proposed in \cite{koreantextoutdoor}. \textbf{CIPED} methods are popular for simple and cost efficient implementation on signboard detection. Applications of CIPED methods can be found in \cite{robust, generic, signboardDANN}. The authors in \cite{generic} showed that their CIPED approach can detect signboards based on a small size dataset containing only 104 images. Another CIPED based system was developed in \cite{robust} where the authors introduced a few image processing techniques such as dilation, erosion and distortion correction to reduce noises. They claimed that their proposed system can detect the signboard area. The recent rise of deep learning based algorithms has inspired researchers to solve the signboard detection problem using \textbf{SBLA} methods \cite{storeclassification, building, signboardDANN}. Generally, SBLA approaches are beneficial for large variety of object detection tasks \cite{surveillance, deepresidual, review, bio2020, detreco, endtoend, VehicleDetection}. The authors in \cite{signboardDANN} showed that the combination of image processing techniques with neural network algorithm can boost the signboard detection and text recognition results. A system was developed using Faster R-CNN (region-based Convolutional Neural Network) algorithm in another research \cite{IFRCNN} where they claimed to have improved the performance of Faster R-CNN on traffic sign detection by introducing secondary RoI proposal method. In \cite{building}, authors proposed a Faster R-CNN based detection algorithm for checking legality of building façades by detecting window, door, advertising objects and signboards. They showed that their method can detect signboards along with other objects. Store classification task was performed in \cite{storeclassification} where they solved the problem by detecting and recognizing store signboards successfully using segmentation based algorithms.

We now discuss the limitations of the existing approaches. Absence of texts or different character orientations and shapes in signboards often causes TSM approaches to fail in detecting the region of the signboards. CIPED methods cannot solve multiple signboard detection problem from image. Such methods do not work well in noisy and overlapping signboard conditions. However, SBLA methods carry the possibility of robust signboard detection system development. Existing SBLA based systems focus on a particular use case of signboard detection for specific country region and these systems cannot differentiate between signboards and other identical objects. Training segmentation based models end to end from scratch degrades model performance, especially when the training dataset is not large enough. If the dataset is different from existing large benchmark segmentation datasets, fine tuning also becomes an obsolete option. Finally, choosing appropriate anchor box dimensions is critical when it comes to SBLA algorithms such as Faster R-CNN. Typical practice is to go with the default values which can be detrimental to dataset specific performance \cite{VehicleDetection}. The authors in \cite{VehicleDetection} have highlighted an anchor box prediction free method based on CNN architecture. So, this method can detect vehicles regardless of perplexing noisy background images by overcoming the limitations of traditional anchor box based method. 

In this research, we provide an extensive method to tackle the limitations of Faster R-CNN based object detection architecture in terms of signboard localization task in densely populated developing cities. Our proposed object detector pipeline consists of a hyperparameter generator algorithm named \textbf{Anchor Ratio \& Anchor Scales (ARAS)} that provides appropriate anchor box dimension to enhance localization accuracy in our algorithm. This algorithm takes into account the objects which are to be localized. The proposed pipeline also includes an automated system for secondary labelled dataset extraction from our original signboard detection dataset. This dataset facilitates pretraining of the convolutional neural network (CNN) backbone that improves our performance further. Finally, we have included an isolated pretraining technique to the proposed pipeline by pretraining only region proposal network portion before moving on to the main training phase, which enhances validation performance.
In response to designing a signboard detection system in the context of developing cities, we introduce a novel dataset named \textbf{Street View Signboard Objects (SVSO)} in this research and currently it is the largest signboard dataset available publicly. Utilizing SVSO annotated dataset, our proposed approach achieves state-of-the-art performance on SVSO dataset and Open Images Dataset V6 (OIDv6) beating contemporary architectures such as Faster R-CNN \cite{FasterRCNN}, YOLOv4 \cite{Yolov4} and YOLOv5 \cite{yolov5}.

In summary, our main contributions are as follows:
\begin{itemize}
\item We design an algorithm named ARAS to generate appropriate proposal box shapes and sizes according to signboard characteristics of the dataset.
\item We have introduced two performance enhancer pretraining techniques - (i) \textbf{SB\_PreCNN} for specialized pretraining of CNN backbone and (ii) \textbf{PreRCNN} for isolated pretraining of region proposal network.
\item Our proposed pipeline for signboard detection has proven to be effective even in scenarios where there are diverse appearances of signboards in a perplexing background.
\item We have provided the largest public signboard dataset with 12,961 signboard objects on 5000 images, which is expected to facilitate further research in this domain.
\end{itemize}

\section{Materials and Methods}
\subsection{Benchmark Dataset Construction}
\label{subsec:dataset}
\begin{figure}
    \centering
    \includegraphics[width=\textwidth]{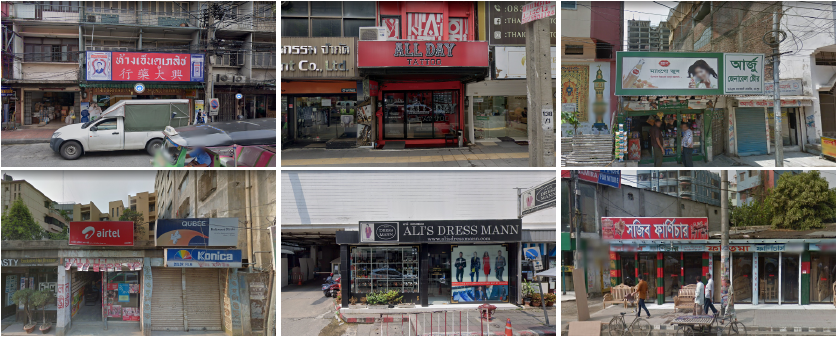}
    \caption{Some sample images from SVSO dataset}
    \label{fig:data}
\end{figure}

We present a novel dataset in this research titled, \textbf{Street View Signboard Objects (SVSO)}, which contains a total of 5000 signboard images of establishments. All images are in PNG file format with bit depth of 32 bits per pixel. The color channels are red, green, blue, and alpha. The SVSO data includes signboard images of various shapes and colours from 10 cities of Bangladesh - Dhaka, Chittagong, Sylhet, Mymensingh, Barisal, Rajshahi, Khulna, Rangpur, Bogra and Pabna. We also constructed an independent test set containing 200 signboard images from 6 developing countries such as Malaysia, Bhutan, Sri Lanka, Myanmar, Thailand and Bangladesh. These natural scene signboard images are labelled in Pascal VOC format and are converted into XML files using an auto labelling tool, labelImg \cite{git}. An annotation file has been generated in CSV format using all the labelled XML files and it contains 12,961 signboard samples with image id, file path, bounding box information (xmin, xmax, ymin, ymax, width) and object class fields. \textbf{Figure \ref{fig:data}} shows some sample images of SVSO dataset. In our dataset collection, all the images were of various sizes (larger than $1000\times600$). Therefore, all image files have been downscaled to $1000\times600$ pixels using Anti-aliasing method. Then image normalization has been performed on these images to transform each pixel value between 0 and 1. To keep all image features on the same scale, standardization method has been applied. SVSO is the largest existing dataset for signboard class objects. We have provided SVSO image dataset, independent test set, labelled XML files, CSV format annotation file and important python scripts in this link: \textit{https://doi.org/10.5281/zenodo.6865241}.

In addition, we have used Billboard class data from Open Images Dataset V6 (OIDv6) \cite{OpenImages}, to analyze the generalizability and effectiveness of our proposed method. The OIDv6 dataset contains around 9M images with a total of 600 object classes. It is one of the largest datasets with 16M object location annotations on 1.9M images. The Billboard class contains a total of 9873 Billboard samples in 4038 images along with bounding box annotation. All Billboard images are in JPG file format with bit depth of 24 bits. 
The Billboard class includes various types of advertisement billboards and establishment signboards with diverse shapes, sizes, colors and background.

\subsection{Faster R-CNN Algorithm Overview for Signboard Detection} \label{subsec:FR-CNN}
\begin{figure}
    \centering
    \includegraphics[width=\textwidth]{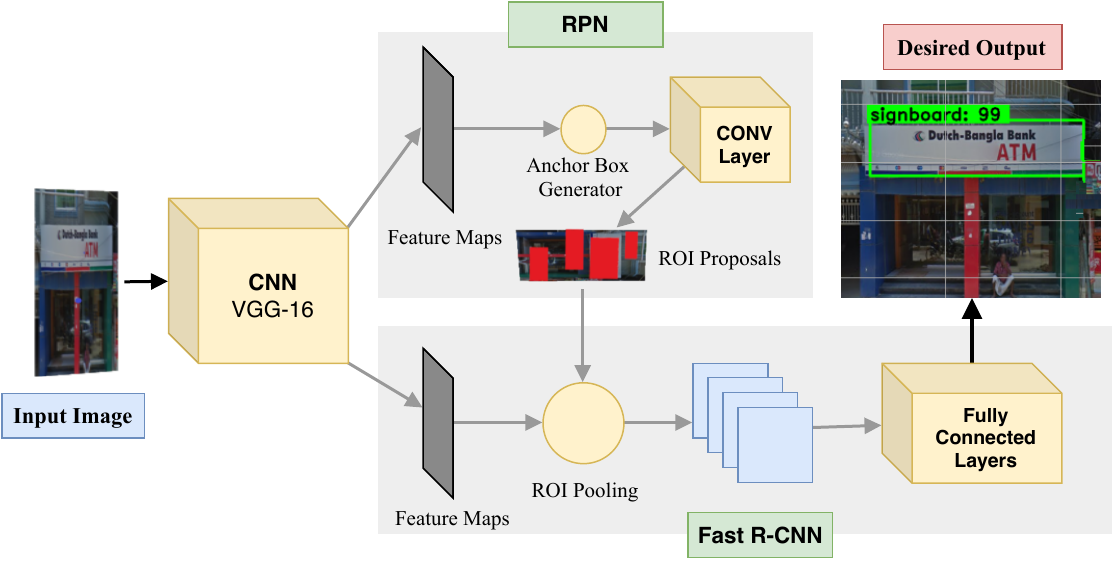}
    \caption{Faster R-CNN architecture: The single unified network for object detection. The RPN module acts as the `attention' and the Fast R-CNN module serves as the `detector' of this architecture.} 
    \label{fig:fr-cnn}
\end{figure}

Our signboard localization approach is based on the workflow of Faster R-CNN which has shown decent performance in various object detection applications \cite{IFRCNN, FRCNNRPN, Pedestrian}. Faster R-CNN can be divided into two independent networks. The first one is region proposal network (RPN) and the second one is fast region based convolutional neural network (Fast R-CNN). The RPN module proposes the regions for possible signboard object locations, while Fast R-CNN module returns the prediction result as object classification and bounding box regression. In \textbf{Figure \ref{fig:fr-cnn}}, we show the unified architecture of Faster R-CNN, which consist of CNN, RPN and Fast R-CNN module. In the following discussion, we briefly illustrate all the main components of Faster R-CNN architecture.

\textbf{CNN} is used in both RPN module and Fast R-CNN detection network as a backbone. Basically, CNN is composed of convolutional layers, ReLU layers and pooling layers. Initially, CNN passes a fixed size input image through these layers for appropriate feature extraction. Then it shares the output feature maps with RPN layer and Fast R-CNN detector to predict the presence of signboard class object in the input image. We bring some improvements in CNN backbone architecture learning process in this research by introducing batch normalization (BN) layers (details in \textbf{Section} \ref{subsec:ourmodel}) and a specialized dataset specific pretraining technique (details in \textbf{Section} \ref{subsec:PreCNN}).

\textbf{RPN} is utilized to find regions which may contain objects by applying the feature matrix extracted by CNN. At first, it takes output feature map from CNN backbone and generates proposal boxes for possible locations of signboard objects in the image. Proposal boxes are produced for each point in the feature map by generating N$\times$M number of anchors, where N and M define the number of predefined anchor scales and anchor ratios, respectively. In the second step, RPN eliminates overlapped proposal boxes using non max suppression (NMS) method and returns the predicted proposal boxes as output. In this research, we propose an algorithm named ARAS for defining appropriate anchor dimensions (details in \textbf{Section} \ref{subsec:ARAS}).

\textbf{Fast R-CNN} consists of RoI pooling layer and fully connected dense layers. Initially, the feature maps from the CNN backbone and the proposal boxes from the RPN layer are sent to RoI pooling layer. Then, RoI pooling layer extracts the features from the backbone feature map using RoI proposals of RPN layer. Generally, max pooling is used to transform these features into fixed size flattened vectors. These vectors from RoI pooling layer are passed into fully connected classifier layers for regression and classification. These layers produce more accurate predictions at every training iteration by minimizing classification and regression losses using softmax and linear activation loss function, respectively. Finally, the network returns the desired detection result by predicting object class type and accurate bounding boxes. We propose isolated pretraining of RPN and Fast R-CNN layers to increase the detection and localization performance of the proposed model (details in \textbf{Section} \ref{subsec:preRCNN}).

\subsection{Pretraining of CNN Backbone} \label{subsec:PreCNN}
Fine tuning approach accelerates the training performance of deep learning model by transferring important information of the pretrained model to the target model \cite{imagenetlarge, imagenet}. Therefore, this approach speeds up convergence of the model \cite{imagenetPTR}. In recent years, many object detection tasks have achieved state-of-the-art results by pretraining with ImageNet dataset \cite{FRCNNRPN,YOLO}. However, ImageNet dataset does not contain any class similar to the objects of signboard class. Besides, images in our dataset contain other objects similar to signboard shape, colour and feature. Therefore, we propose a pretraining method named \textbf{Signboard PreCNN (SB\_PreCNN)} to fine tune the CNN backbone.

\subsubsection{Pretrained Model Preparation}
Our approach is based on pretraining CNN backbone architecture on signboard and non-signboard images. To prepare this pretrained model, we build another secondary training dataset from the original training set images of SVSO dataset using \textbf{Object Region Extractor (ORE)} method (described in \textbf{Section} \ref{subsec:rbe}). This dataset contains total 20000 images of signboard class and non-signboard class (class balanced), and the shape of each image is 224$\times$224$\times$3. At the beginning of SB\_PreCNN model training, we pass the newly created signboard and non-signboard images through convolutional and pooling layers of the CNN architecture to filter out signboard and non-signboard features. Finally, we send the feature maps to the fully connected layer and compute the binary class probability scores using Sigmoid function. We train this model up to 60 epochs and further use it for fine tuning during our main training phase. 

\subsubsection{Object Region Extractor} \label{subsec:rbe}
\begin{figure}
    \centering
    \includegraphics[width=0.8\linewidth]{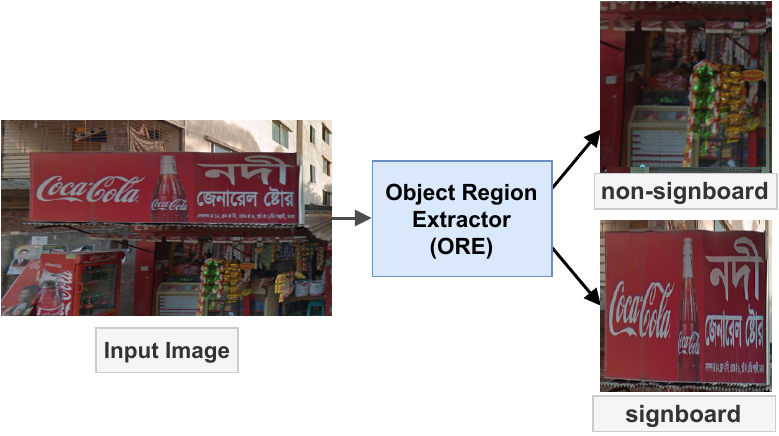}
    \caption{Signboard and non-signboard image generation using Object Region Extractor (ORE) method}\label{fig:ORE}  
\end{figure}

\textbf{Object Region Extractor (ORE)} method extracts each class objects (signboard) and background (non-signboard) regions from an image without the need of any additional manual labeling. This algorithm takes an image and its signboard labeling details (manual bounding box annotation for signboard localization model training) as input. Then, we extend labelled bounding box range by 10 pixels on both dimensions so that we can crop signboard regions appropriately. To build signboard class images, we just crop out the labelled bounding box range from each image using corresponding xmin, xmax, ymin and ymax values. After that, we collect non-signboard images using sliding window method. Here, size of each window is $224\times224$ and we slide every window position on an image by 1 pixel, so that we can verify whether the covered area is overlapped with any labelled bounding box range or not. If a sliding window block does not belong to any labelled bounding box range of signboard class, it indicates that current window block is on non-signboard portion of the image. And then we crop out this block from the image and save it as non-signboard image. Using ORE method, we have created 20000 signboard and non-signboard images. In \textbf{Figure \ref{fig:ORE}}, we show a sample signboard \& non-signboard class generation from an original image based on ORE method.

\subsection{Pretraining of RPN and Fast R-CNN Detection Network} \label{subsec:preRCNN}
Generally, RPN and Fast R-CNN layers are initialized with randomly generated weights and are placed on top of CNN backbone layers, where CNN layers are initialized with pretrained weights \cite{IFRCNN, FRCNNRPN}. However, in terms of training speed and performance, pretraining approaches give better performance than random initialization \cite{surfacedefect}. We propose a pretraining technique named \textbf{PreRCNN} for RPN and Fast R-CNN detection network. We prepare another pretrained model by training only RPN layer and Fast R-CNN layer of Faster R-CNN architecture on SVSO dataset, while making other layers non-trainable (frozen). We save the updated weights and use them for fine tuning during the main training phase. Such isolation training helps these detection layers get a solid grasp over the dataset unique characteristics.

\subsection{Appropriate Anchor Scale and Anchor Ratio Determination} \label{subsec:ARAS}
\begin{figure}
    \centering
    \includegraphics[width=0.8\linewidth]{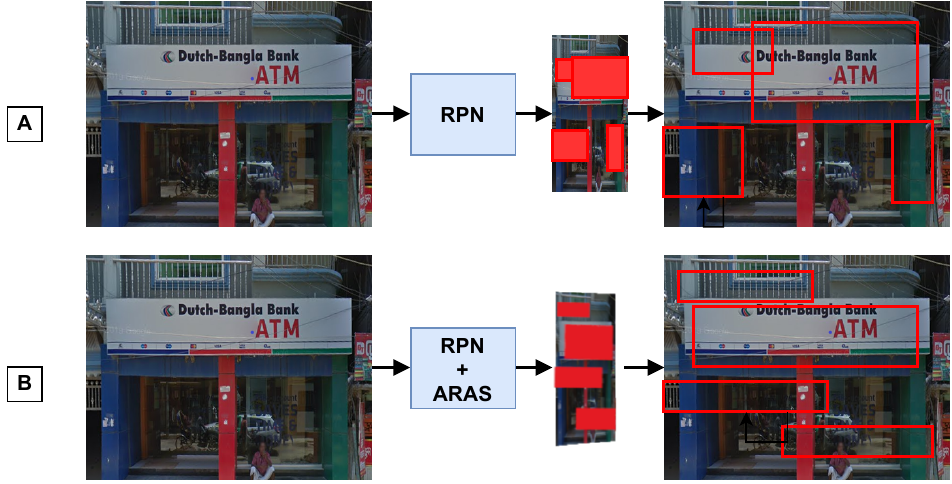}
    \caption{Generation of RPN layer anchor boxes: (A) - ROI proposals based on default anchor ratios and scales. (B) - ROI proposals based on proposed ARAS algorithm.}\label{fig:PRNar}  
\end{figure}

Anchor box generation is a widely used technique in state-of-the-art localization algorithms such as Faster R-CNN, YOLO etc. \cite{FRCNNRPN, YOLO}. In Faster R-CNN, these anchor boxes are generated by RPN layer using pre defined anchor ratios and anchor scales. Initially, we started our experiments according to the default anchor box dimensions provided in \cite{FRCNNRPN}. However, we have found out that without appropriate anchor ratios and anchor scales in accordance with signboard shapes of the dataset, we cannot lead the model towards our desired performance. Therefore, we propose an algorithm named \textbf{Anchor Ratio \& Anchor Scales (ARAS)} to select appropriate anchor box dimensions. \textbf{ARAS} algorithm is completely automatic and needs no additional manual processing. As part of labeling for any localization algorithm, we need to label ground truth bounding boxes in training set images. The proposed ARAS algorithm takes these bounding boxes as its input and does not require any other additional labeling. It generates appropriate anchor ratios and scales of class objects. \textbf{Figure \ref{fig:PRNar}} shows the RPN layer's anchor box suggestions before and after using ARAS algorithm. It is clear that RPN layer anchor box suggestions become much more appropriate when ARAS determined anchor ratios and anchor scales are used. We illustrate the algorithm steps as follows.

\begin{figure}[h]
    \centering
    \includegraphics[width=0.65\linewidth]{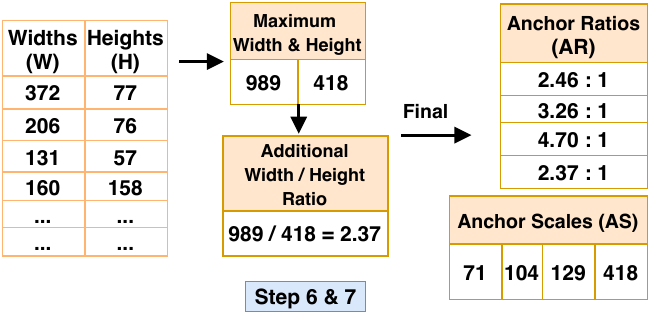}
    \caption{\textbf{Step 6 \& 7} ARAS algorithm: final anchor ratios (AR) and anchor scales (AS) are determined by including max aspect ratio and max height}\label{fig:stp6-7}
\end{figure}
\begin{figure}[h]
    \centering
    \includegraphics[width=0.55\linewidth]{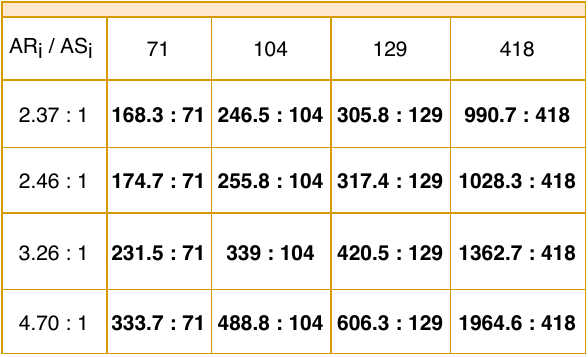}
    \caption{Anchor box sizes and shapes according to anchor ratios (AR) and anchor scales (AS), which are generated from ARAS algorithm.}\label{fig:aras}
\end{figure}

\begin{enumerate} 
    \item \textbf{Figure \ref{fig:stp1-3}} illustrates Step 1 to 3. At first, we create a data table to store all signboard width and height values from our dataset, which we use as an input of our ARAS algorithm.
    \item Then, we read width and height values of signboard from the data table and create a separate list for widths and heights denoted as $W$ and $H$, respectively.
    \item Since scale and ratio are the major features for accurate anchor box generation, a proper clustering algorithm is necessary for generalized feature extraction \cite{Survey,locallearning,micro}. Therefore in this step, we determine $K$ number of width and height centers using the K-Means clustering algorithm, where $K = 3$. The value of $K$ indicates number of anchor ratio and anchor scales that are available at each feature cell during anchor box proposal generation. After experimenting with different values of $K$ (clustering with $K = 3$ and $K = 4$, we observed validation $mAP$ score of 0.91 and 0.85, respectively), we find $K=3$ to give the best validation performance. We have found out that larger values of $K$ cause overfitting of the model by feeding extensive amount of proposals during RPN training. Also in \cite{FasterRCNN}, the authors report 3 scales with 3 aspect ratios to deliver better generalization in object detection. Now the clustered width and height centers are denoted as $WC$ and $HC$, respectively. $WC$ and $HC$ values are important for aspect ratio calculation. 
    
\begin{figure}[h]
    \centering
    \includegraphics[width=\linewidth]{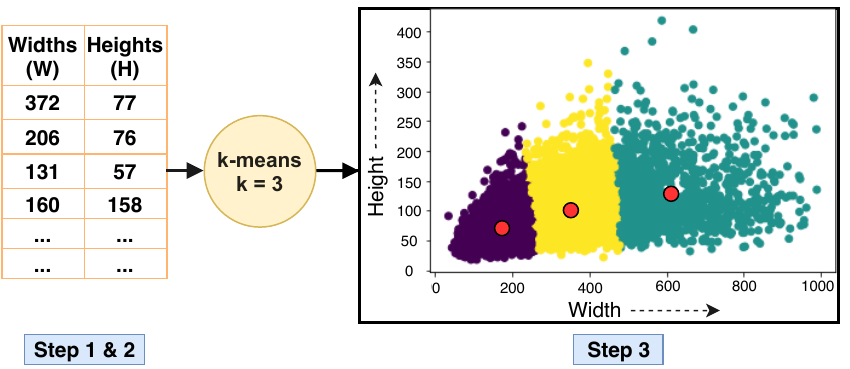}
    \caption{\textbf{Step 1-3} of ARAS algorithm: 3 center values for widths and heights are calculated using K-Means clustering.}
    \label{fig:stp1-3}
\end{figure}
\begin{figure}[h]
    \centering
    \includegraphics[width=0.7\linewidth]{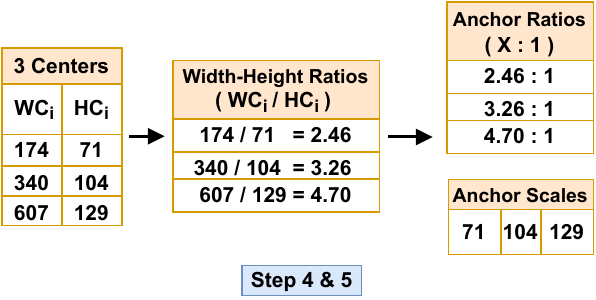}
    \caption{\textbf{Step 4 \& 5} of ARAS algorithm: using width centers ($WC$) and height centers ($HC$), anchor ratios (AR) and anchor scales (AS) are determined}
    \label{fig:stp4-5}
\end{figure}

    \item \textbf{Figure \ref{fig:stp4-5}} illustrates Step 4 and step 5. We calculate our desired aspect ratios ($X:Y$) using $WC$ and $HC$ values. We divide each width center, $WC_i$, by each height center, $HC_i$. Then we set $X_i = WC_i/HC_i$ and $Y_i=1$ (each aspect ratio is in $X:1$ format). Let us understand the significance of this aspect ratio. Suppose, $HC_i$ = 100 (scale) and $X_i$ = 2.5 (ratio). Then the width and height for the generated anchor box will be $100 \times 2.5=250$ and $100 \times 1=100$, respectively.  

    \item As discussed above, appropriate proposal boxes can be produced by multiplying each height center $HC_i$ with corresponding aspect ratio $X_i: 1$, where the box width and height will be $HC_i \times X_i$ and $HC_i \times 1$, respectively. So, until now we have calculated $K$ number of anchor ratios ($X_i$) and anchor scales ($H_i$). 
    
    \item \textbf{Figure \ref{fig:stp6-7}} shows step 6 and 7. At this phase, we generate maximum width $(W_{max})$ and maximum height $(H_{max})$ from width ($W$) and height ($H$) list of Step 1. Then, we calculate additional Anchor Ratio by dividing $W_{max}$ by $H_{max}$ and pick $H_{max}$ as additional Anchor Scale. This step is necessary for handling the cases of large signboards, as K-Means clustering faces challenging issues while clustering large valued outlier data points of the dataset \cite{locallearning}. So, if we choose $K=3$, ultimately we end up generating total 4 anchor ratios and 4 anchor boxes.
    
    \item Finally, we get $K+1$ number of Anchor Ratios (AR) and Anchor Scales (AS) by including the maximum ratio and the maximum scale, respectively. \textbf{Figure \ref{fig:aras}} shows the proposal box sizes and shapes generated from ARAS algorithm.
\end{enumerate}

\begin{figure}[h]
   \centering
    \includegraphics[width=0.65\linewidth]{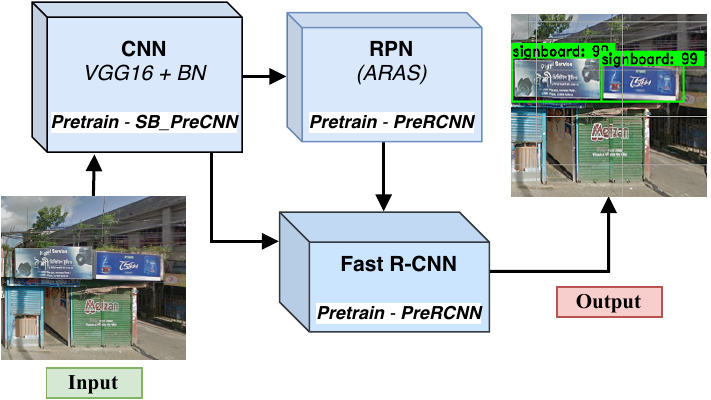}
    \caption{Unified architecture of our proposed object detection algorithm.}
    \label{fig:model}
\end{figure}

\subsection{Integration of the Proposed Methodology} \label{subsec:ourmodel}
 \textbf{Figure} \ref{fig:model} shows our final signboard detection model architecture. Initially, we have designed the CNN backbone with an improved version of VGG16 architecture, where, Batch Normalization (BN) has been added between each convolution and activation layer. BN reduces the hidden layer unit value covariance shift and it resists overfitting of the model. Besides, it normalizes the previous layer's output and accelerates the model performance \cite{imagenet}. We have improved RPN's anchor box generation performance by determining anchor ratios and anchor scales using ARAS algorithms. Finally, during our model construction, we have pretrained CNN backbone and RPN \& R-CNN network based on SB\_PreCNN and PreRCNN pretraining techniques, respectively before the actual training. 

\section{Results and Discussion}

\subsection{Experimental Setup}
\label{subsec:setup}

In this research, we have used Google Colab \cite{Bisong2019} for all the experiments which is a cloud based server that give access to powerful GPU runtime environments. We utilize two such separate Google Colab servers simultaneously that include both Tesla p4 and Tesla P100 16GB GPU environments.

We conduct experiments on both SVSO and OIDv6 datasets. We use 4000 images on the training set and 1000 images on the validation set from SVSO signboard dataset. Moreover, we use 200 images from cities of six different countries as independent test set. On the other hand, from Billboard class of OIDv6 dataset, we use 3230 images on the training set and 808 images on the test set. No hyperparameter tuning was done based on any portion of the OIDv6 dataset.

We train our unified architecture (discussed in \textbf{Section} \ref{subsec:ourmodel}) for 100 epochs with early stopping condition and with a learning rate of 0.0001. We have used batch size of 512. We use ARAS algorithm generated aspect ratios (2.37:1, 2.46:1, 3.26:1, 4.70:1) and scales (71, 104, 129, 418) for anchor box ratios and anchor box scales hyperparameter, respectively. For activation function, we use ReLU in hidden layers of CNN backbone, Sigmoid activation in RPN layer and Softmax activation in the final classifier layer. We utilize Stochastic Gradient Descent (SGD) optimizer for weight update during training. We have chosen Mean Absolute Error (MAE) as our loss function. We comprehensively evaluate the proposed model based on $mAP$ score where Intersection Over Union $(IoU)$ threshold has been set to 0.7. The details of our experiments and corresponding results have been described in the following subsection.

\subsection{Experiments and Results}
\label{subsec:res}

We have compared the detection performance of our proposed algorithm with baseline methods based on $mAP$ (mean average precision) score. Calculation of Intersection Over Union $(IoU)$, $Precision$ and $Recall$ are required to determine the $mAP$ score. $IoU$ determines a score from 0 to 1 that indicates the range of overlap between the predicted and ground truth bounding boxes. 
\[ IoU = \frac{area \; of \; intersection \; of \; ground \text{-} truth \; and \; predicted \; boxes}{area \; of \; union \; of \; ground \text{-} truth \; and \; predicted \; boxes} \]
We consider a predicted signboard bounding box to be True Positive (TP) only if $IoU$ value of that box with a ground truth bounding box is greater than or equal to 0.7. Otherwise, we consider the prediction as False Positive (FP). Here, $Precision$ is the ratio of TP and the total number of predicted positives (TP+FP), while $Recall$ is defined as the ratio of TP and total number of ground truth bounding boxes (TP+FN).
\[ Precision = \frac{TP}{TP+FP} \]
\[ Recall = \frac{TP}{TP+FN} \]
So the $mAP$ is mean of Average Precision (AP) over N object classes and the AP is calculated from the area under the $Precision$ vs $Recall$ curve.
\[ mAP = \frac{\sum_{i=1}^{N} AP_{i}}{N} \]

All experiments have been performed using 80\%-20\% training-validation split. We start with the default hyperparameters (D Param.) of originally proposed Faster R-CNN \cite{FRCNNRPN} which are as follows: 9 anchor boxes with anchor scales of ($128$, $256$, $512$) and anchor ratios of (1:1, 1:2, 2:1), IoU threshold of 0.7 for NMS and maximum box suppression of 300 for NMS output. We also initially use ImageNet pretrained CNN backbone fine tuning like most relevant researches. We show step by step evaluation and comparison starting from this baseline in order to display the effectiveness of our proposed approach in signboard localization. 

Initially, we experiment with different set of processed images of SVSO dataset using default hyperparameters. In this regard, three different variations of images have been prepared - (i) RGB image data have been resized using OPENCV INTER AREA method, (ii) GRAY is a gray scale image obtained using OPENCV BGR2GRAY method and (iii) PIL\_RGB has been resized using PIL ANTIALIAS method \cite{opencv_library}. We have achieved the best performance on PIL\_RGB image dataset. Generally, resizing images using PIL ANTIALIAS process retains most of the important features of the original image, however it requires higher computational time. We choose PIL\_RGB formation of SVSO dataset for further experiments. Relevant performances are provided in \textbf{Table} \ref{da}.

\begin{table}[h]
  \begin{center}
  \textbf{\caption{Faster R-CNN performance on validation set for application of different image preprocessing techniques. VGG16 has been used as CNN backbone. \strut\strut\label{da}}}
    \vspace*{1mm}
    \resizebox{0.4\textwidth}{!}{%
    \begin{tabular}{|c|c|}
   \hline
   \textbf{Dataset Variant} & \textbf{mAP}\\
   \hline
   GRAY & 0.65\\
   \hline
   RGB & 0.70\\
   \hline
   PIL\_RGB & 0.73\\
   \hline
  \end{tabular}}
  \end{center}
\end{table}

In the second phase, we evaluate different CNN architectures \cite{deepresidual, smalltraining, densely, inceptionresnet}. VGG16 architecture has shown comparatively better performance than InceptionV4, Resnet50, Densenet121 and Densenet201. Depth of VGG16 architecture and its simple sequential nature prove to act as effective feature extractor for signboard localization. We use VGG16 in all further experiments as our Faster R-CNN backbone architecture. The results for CNN architecture experiments are shown in \textbf{Table} \ref{BA}.

\begin{table}[h]
  \begin{center}
  \textbf{\caption{Experimental results of different CNN backbones (fine tuned from ImageNet pretraining) while using default hyperparameters and PIL\_RGB image processing technique on SVSO validation set. \strut\strut\label{BA}}}
    \vspace*{1mm}
    \resizebox{0.4\textwidth}{!}{%
    \begin{tabular}{|c|c|}
   \hline
   \textbf{CNN Backbone} & \textbf{mAP}\\
   \hline
   InceptionV4 & 0.56\\
   \hline
   Resnet50 & 0.58\\
   \hline
   Densenet121 & 0.61\\
   \hline
   Densenet201 & 0.64\\
   \hline
   VGG16 & 0.73\\
   \hline
  \end{tabular}}
  \end{center}
\end{table}

\begin{table}[h]
  \begin{center}
  \textbf{\caption{Performance comparison of our proposed final model with baseline methods on SVSO validation set in terms of $mAP$ score \label{PM}}}
    \resizebox{0.9\textwidth}{!}{%
    \begin{tabular}{|p{2cm}|p{2cm}|p{3.1cm}|p{3cm}|p{1cm}|}
    \hline
    \textbf{Backbone CNN} & \textbf{CNN Pretraining} & \textbf{RPN \& Fast R-CNN Pretraining} & \textbf{Anchor Ratio \& Anchor Scale} & \textbf{mAP}\\
    \hline
    VGG16 & - & - & D Param. & 0.69\\
    \hline
    VGG16 & ImageNet & - & D Param. & 0.73\\
    \hline
    VGG16 & SB\_PreCNN & - & D Param. & 0.78\\
    \hline
    VGG16+BN & SB\_PreCNN & - & D Param. & 0.81\\
    \hline
    VGG16+BN & SB\_PreCNN & PreRCNN & D Param. & 0.85\\
    \hline
    VGG16 & SB\_PreCNN & PreRCNN & ARAS Algorithm & 0.88\\
    \hline
    VGG16+BN & SB\_PreCNN & PreRCNN & ARAS Algorithm & 0.91\\
    \hline
    \end{tabular}}
  \end{center}
\end{table}

\begin{figure}[h]
    \centering
    \includegraphics[width=0.85\linewidth]{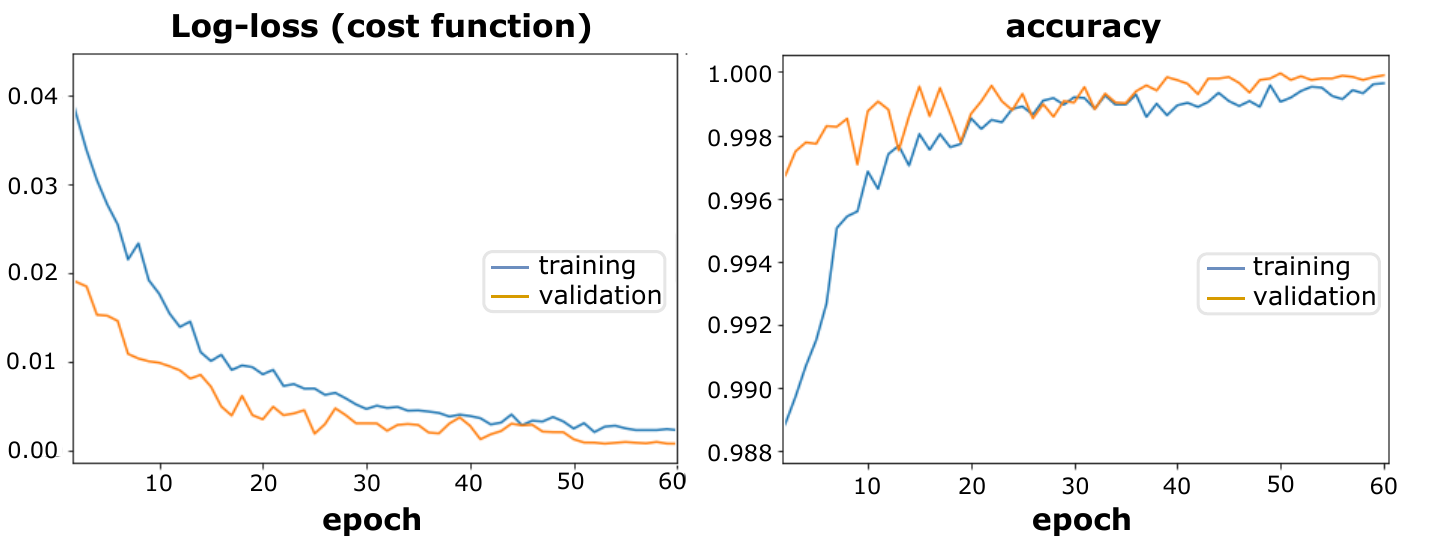}
    \caption{Training and validation loss-accuracy graph of VGG16 CNN model pretraining which has been used later as CNN backbone of Faster R-CNN} \label{fig:preCNN}
\end{figure}

In the third stage, we experiment on CNN backbone pretraining technique. Our proposed pretraining technique SB\_PreCNN (training based on signboard and non-signboard image classification) showed improved performance on selected VGG16 architecture. \textbf{Figure \ref{fig:preCNN}} shows training and validation loss \& accuracy graph of CNN backbone pretrained model on the signboard and non-signboard classification dataset. The $1st$ row of \textbf{Table \ref{PM}} shows the experiment result of training from random state. The results of pretraining experiments of ImageNet and SB\_PreCNN are shown in the $2^{nd}$ and $3^{rd}$ row of \textbf{Table \ref{PM}}, respectively. The reason behind achieving better result using SB\_PreCNN pretraining is that it helps the CNN backbone to learn distinguishing features between signboard and non-signboard object before start of the mainstream training. It is to note that ImageNet dataset does not contain any class matching signboard type objects.

In the fourth step, we add batch normalization (BN) between each convolution and activation layer of VGG16 architecture used as CNN backbone. The $4th$ row of \textbf{Table \ref{PM}} shows the improved performance after this addition. Batch normalization helps to prevent overfitting of the model and to increase training stability.

\begin{figure}[h]
    \centering
    \includegraphics[width=0.4\linewidth]{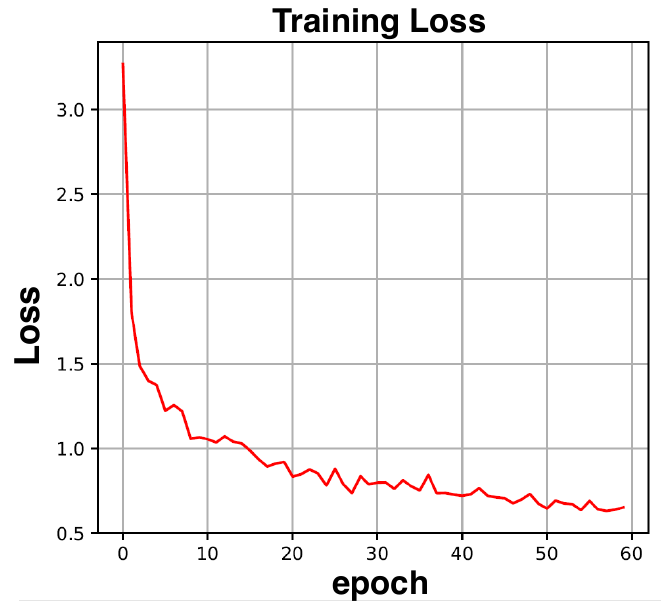}
    \caption{Training loss graph of the RPN \& Fast R-CNN layer's pretrained model.} \label{fig:RCNN}
\end{figure}

In phase number five, we add PreRCNN pretraining scheme which has geared up the training performance of our model shown in the $5th$ row of \textbf{Table \ref{PM}}. PreRCNN is our proposed pretraining technique for RPN \& Fast R-CNN layer. \textbf{Figure \ref{fig:RCNN}} shows training loss graph of this pretraining. Isolated pretraining of the mentioned layers help them get an idea of appropriate region proposal and localization independent of CNN backbone feature extraction quality, which helps faster convergence during end to end training. 

\begin{figure}[h]
    \centering
    \includegraphics[width=0.9\textwidth]{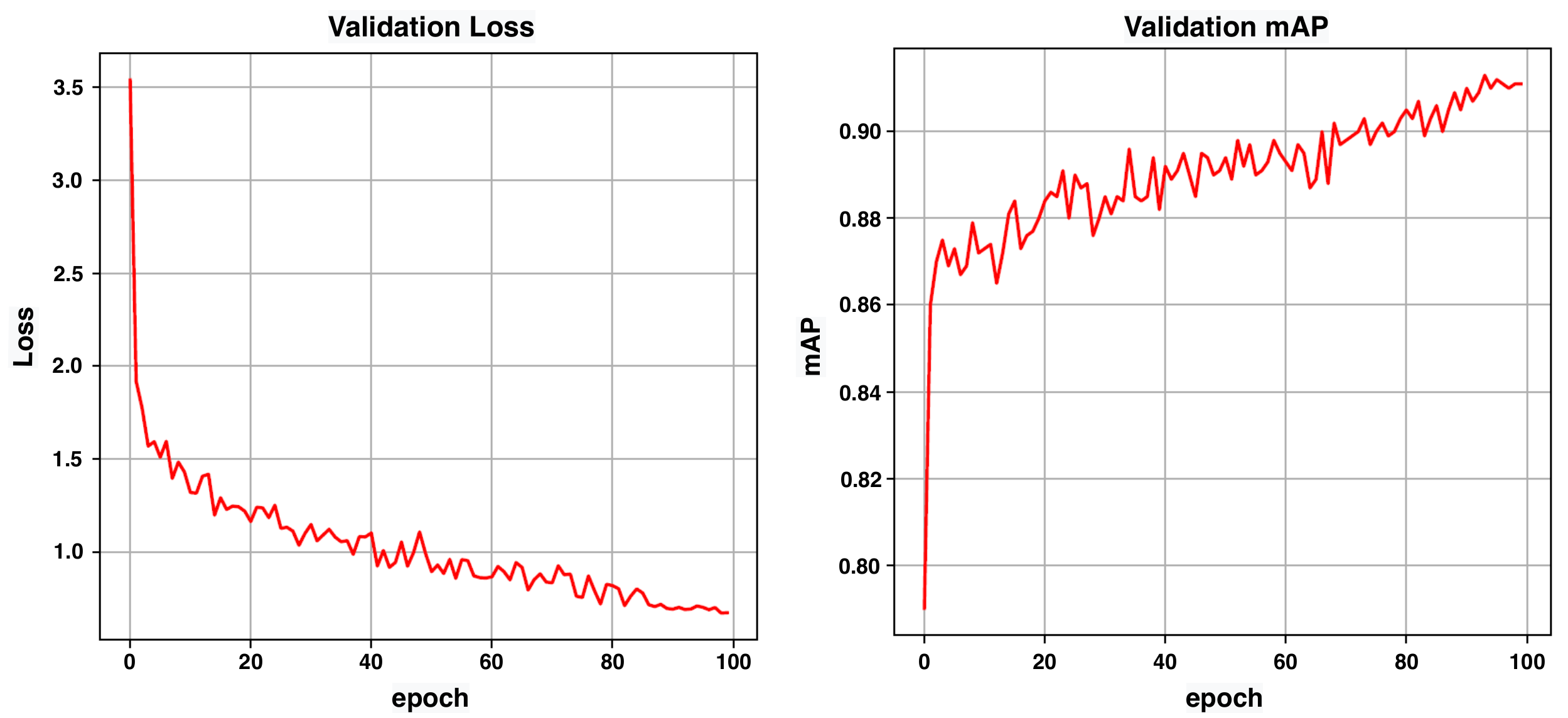}
    \caption{Validation loss and $mAP$ score graph of our proposed algorithm mainstream training.}
    \label{fig:res}
\end{figure}

Finally, we use the anchor ratios (2.37:1, 2.46:1, 3.26:1, 4.70:1) and anchor scales (71, 104, 129, 418) generated by our proposed ARAS algorithm instead of default values. Choice of these hyperparameters have provided us with substantial validation performance improvement shown in the $7th$ row of \textbf{Table \ref{PM}}. ARAS algorithm helps provide near optimal RoI proposals in accordance to the dataset. We thus finish our experiments with an impressive validation $mAP$ score of 0.91. \textbf{Figure} \ref{fig:res} depicts the validation loss and $mAP$ score graph of our proposed algorithm mainstream training. Among the modifications that we have made, introduction of batch normalization is perhaps the simplest. As part of ablation study, we remove batch normalization from our final proposed model to see the performance difference (see row 6 of \textbf{Table \ref{PM}}). We see a drop in model performance without batch normalization.

\begin{figure}[h]
    \centering
    \includegraphics[width=0.7\textwidth]{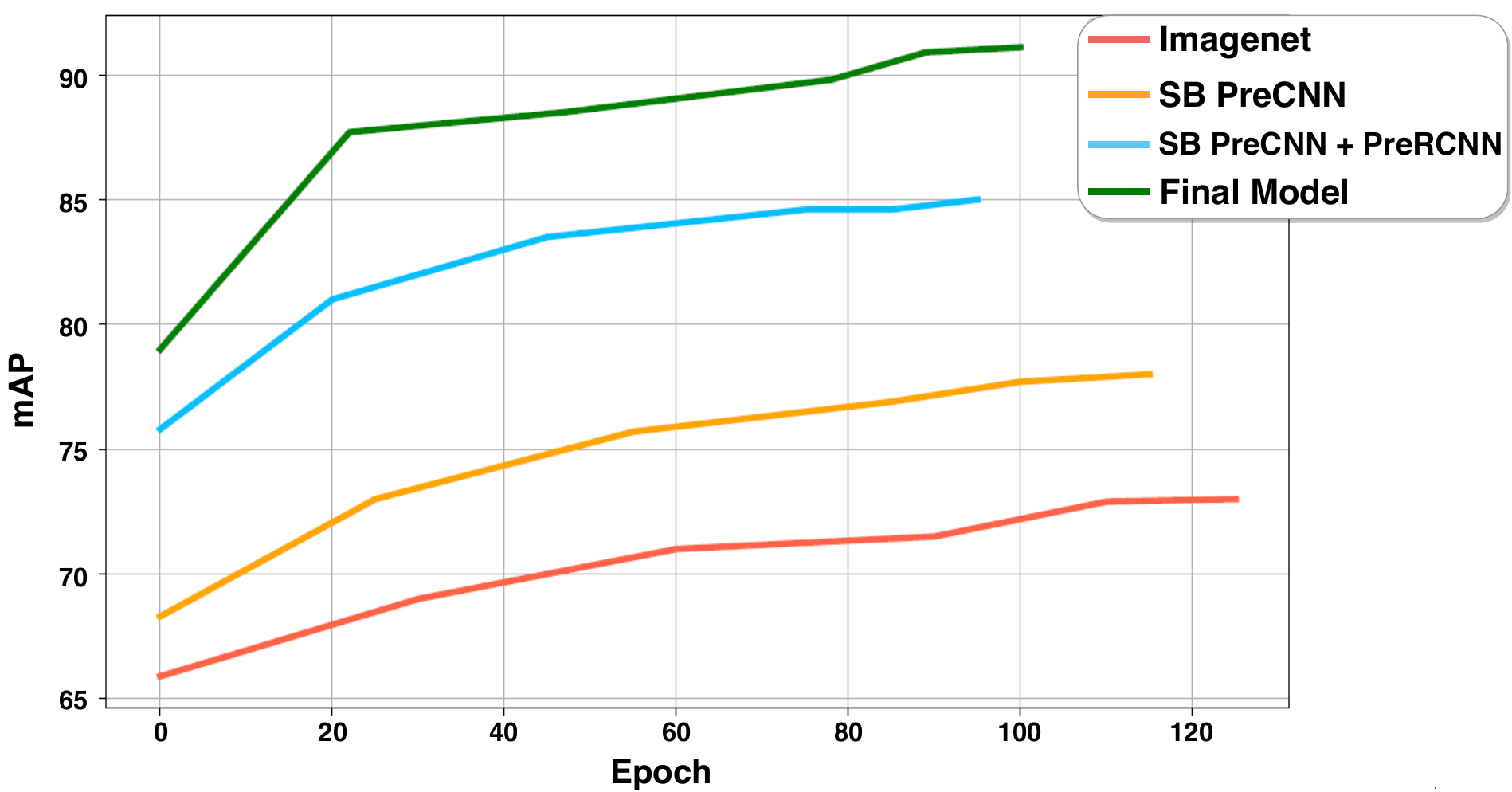}
    \caption{Validation $mAP$ score vs training epoch number curves: comparison after inclusion of proposed modifications}
    \label{fig:tradeoff1}
\end{figure}

In \textbf{Figure \ref{fig:tradeoff1}}, we have shown number of training epochs vs validation $mAP$ score curves of different stages of Faster R-CNN localization algorithm modification. We can see clear improvement in terms of performance and convergence of our proposed final model from this comparison plot.  



In summary, our final model includes the following modifications:
\begin{itemize}
    \item Batch normalized VGG16 as CNN backbone
    \item Proposed SB\_PreCNN pretraining technique applied on the CNN backbone
    \item Proposed PreRCNN isolation pretraining applied on RPN and Fast R-CNN layers
    \item Anchor ratios and anchor scales generated by implementing ARAS algorithm
\end{itemize}

In the following discussion, we present several experiments that exhibit effectiveness and generalization ability of the proposed method. These experiments include comparison with more advanced methods, analyzing robustness of our model and testing the proposed method on several datasets.

\begin{figure}[h]
    \centering
    \includegraphics[width=0.7\textwidth]{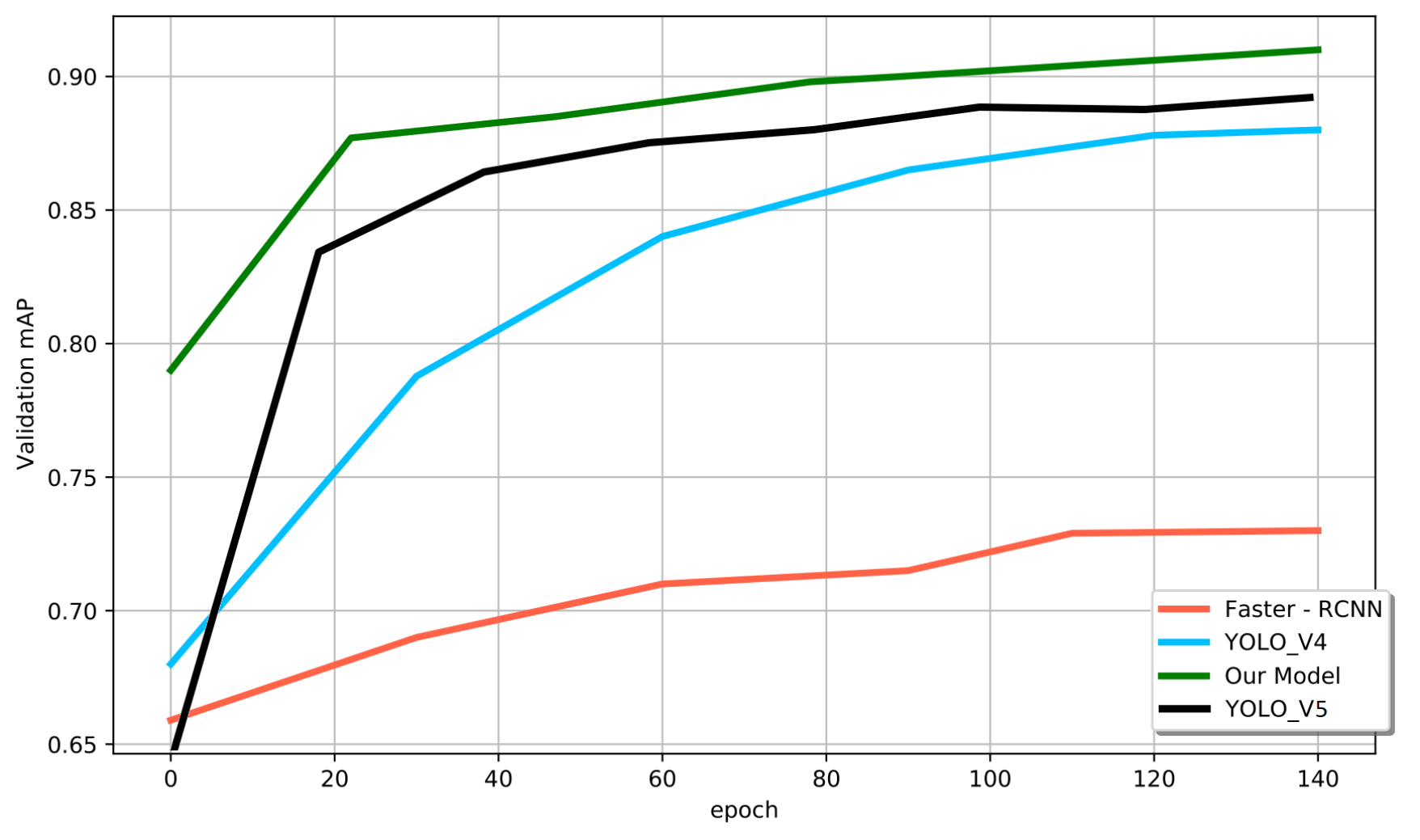}
    \caption{Validation $mAP$ score vs training epoch number curves: comparative analysis of our proposed model with current state-of-the-art methods}
    \label{fig:sota}
\end{figure}

We have compared our proposed signboard localization algorithm with advanced localization algorithms such as YOLO \cite{YOLO}, Faster R-CNN \cite{FasterRCNN}, YOLOv4 \cite{Yolov4} and recent state-of-the-art detection method YOLOv5 \cite{yolov5, Yolov5Comp} on SVSO dataset based on validation $mAP$ score. The scores have been provided in Table \ref{SOTA}. Although YOLO has fast validation speed, it shows comparatively poor result on SVSO because of its over simplistic architecture design based on mostly mean squared error type loss functions and an end to end trainable CNN architecture. Since our motivation behind this research is to be able to localize signboards from static images obtained from longitude latitude marked repositories such as Google Street View, real time signboard localization is not desired here. Rather, we try to maximize performance in a reasonable amount of validation time. Epoch by epoch validation $mAP$ scores obtained while training on SVSO dataset of Faster R-CNN, YOLOv4, YOLOv5 and our proposed method have been provided in Figure \ref{fig:sota}. 

\begin{table}[h]
  \begin{center}
  \textbf{\caption{Comparative result of our proposed algorithm with current state-of-the-art methods on SVSO validation set \label{SOTA}}}
    \vspace*{1mm}
    \resizebox{0.4\textwidth}{!}{%
    \begin{tabular}{|c|c|}
   \hline
   \textbf{Localization Algorithm} & \textbf{mAP}\\
   \hline
   YOLO & 0.66\\
   \hline
   Faster R-CNN & 0.73\\
   \hline
   YOLOv4 & 0.88\\
   \hline
   YOLOv5 & 0.89\\
   \hline
   Proposed Algorithm & 0.91\\
   \hline
  \end{tabular}}
  \end{center}
\end{table}

YOLOv4 and YOLOv5 both show close performance compared to our proposed method. YOLOv4 is based on two types of YOLO architecture modifications - bag of freebies which occur no extra inference cost and bag of specials which affect the inference cost. Introduction of CIoU (Complete IoU) loss, Self-adversarial training and simple K-means clustering based anchor box selection are examples of introduced bag of freebies. On the other hand, Mish activation function usage, using CSPDarkNet (Cross Stage Partial Network in Darknet) as CNN backbone, Spatial Pyramid Pooling (SPP), Path Aggregation Network (PANet), Spatial Attention Module (SAM) and multi-resolution output prediction are examples of bag of specials introduced in YOLOv4 architecture. Compared to these modifications in base YOLO, our proposed modifications in base Faster R-CNN architecture are simple and incur no additional inference cost. Moreover, these modifications can be applied to other localization algorithms as well to improve their dataset specific performance. On the contrary, YOLOv5 is lighter compared to previous YOLO versions. It's comparatively more advanced and faster than YOLOv4 and YOLOv3 due to its newly introduced CNN backbone \cite{Yolov5Comp}. YOLOv5 uses a unified structure of CSPdarknet53 with a Focus layer as CNN backbone. The major benefits of a Focus layer are - it decreases CUDA memory usage, eliminates initial 3 layers of YOLOv3 by replacing with a single layer, and enhances both forward propagation and backpropagation. YOLOv5 uses only PANet as neck, it extracts feature pyramids and enhances the localization accuracy by increasing flow of low level features in the model.

\begin{figure}[!htb]
 \centering
     \subfloat{\includegraphics[width=1.61in]{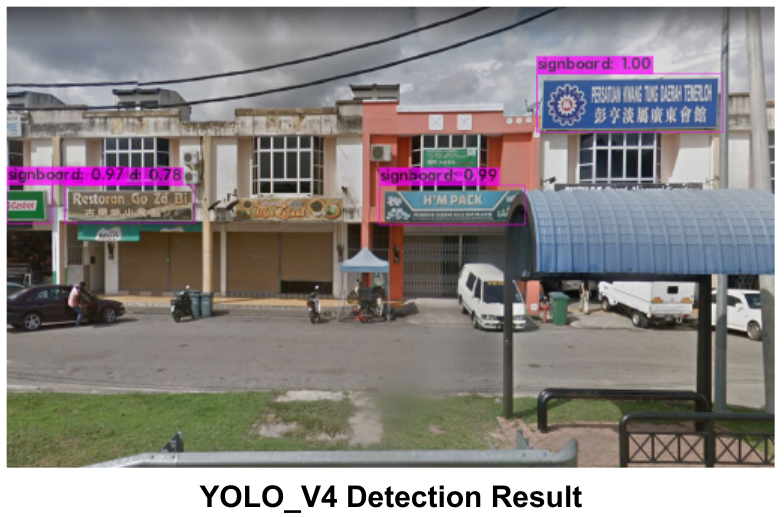}}
     \subfloat{\includegraphics[width=1.6in]{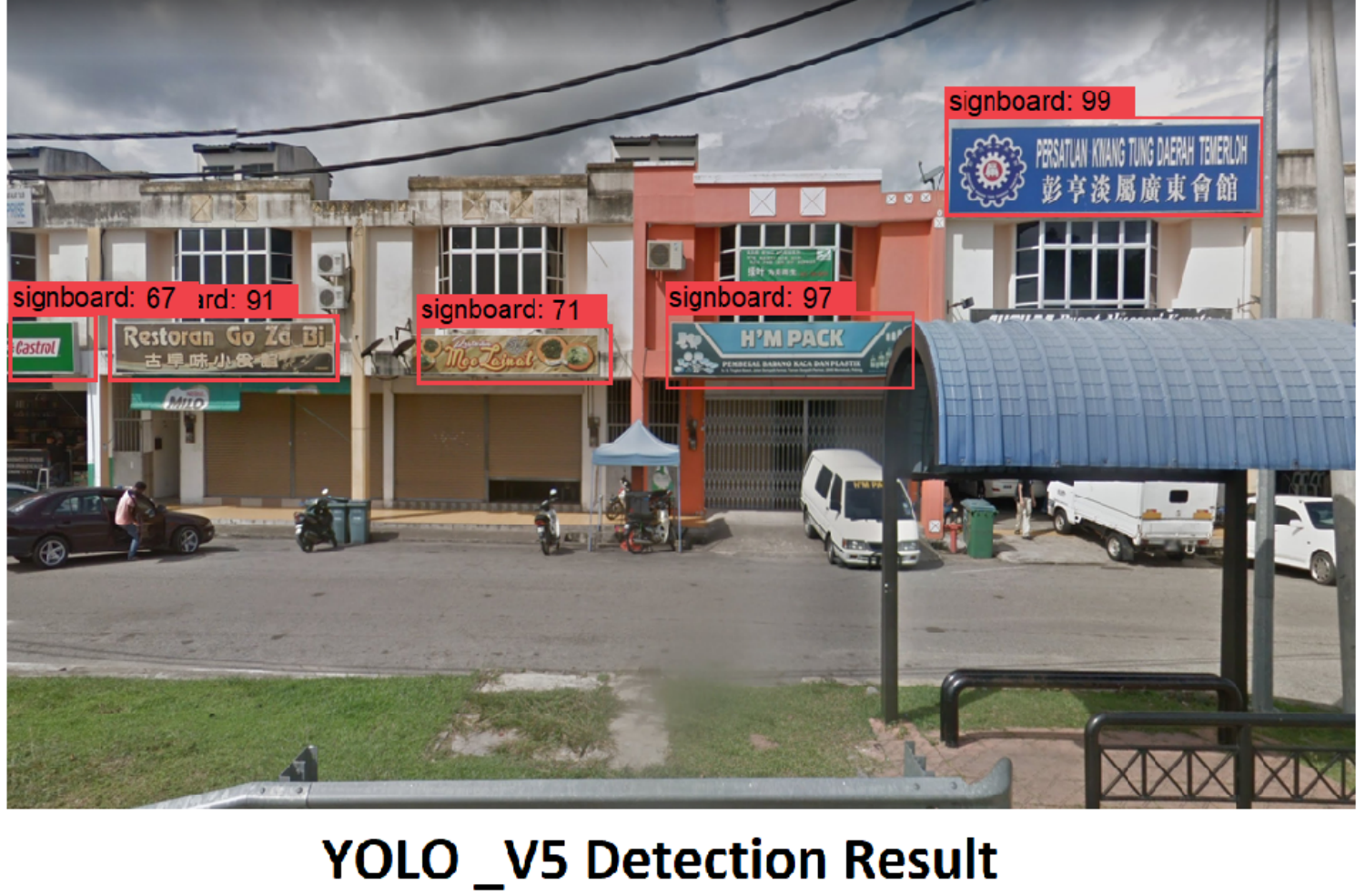}}
     \subfloat{\includegraphics[width=1.6in]{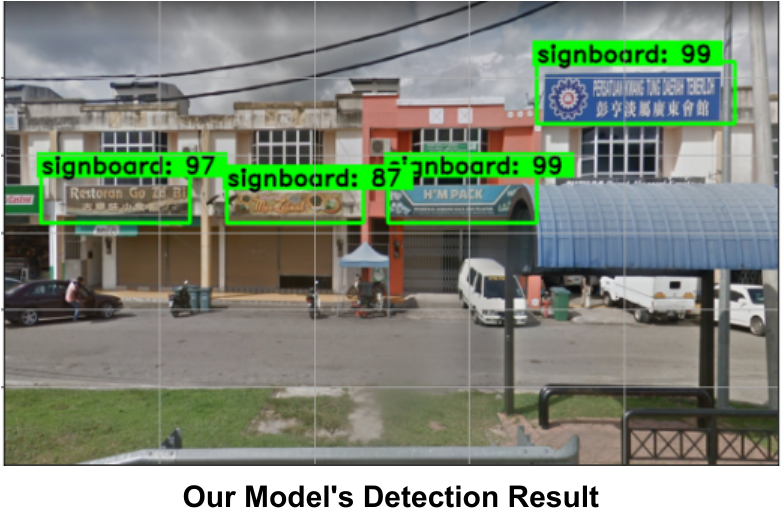}} 
\caption{Detection performance of YOLOv4, YOLOv5, and our proposed algorithm on a sample SVSO test set image}
\label{fig: image_comp}
\end{figure}

Figure \ref{fig: image_comp} shows detection and localization performance of YOLOv4, YOLOv5 and our proposed algorithm on a sample SVSO test set image. Here YOLOv4 misses one signboard in this image and YOLOv5 detects signboards with a lower confidence score compared to the detection result of our proposed algorithm. 

We now test the robustness of our method through undersampling our training set. For training, we use only 10\% of SVSO dataset training images (500 images). We then evaluate the model using our validation set which has given us an $mAP$ score of 0.84 (0.91 $mAP$ score when using the full training set). Although our approach is Faster R-CNN based, Faster R-CNN gives only 0.69 validation $mAP$ score in spite of using the entire training set. This shows that our method keeps its accuracy even when the dataset is comparatively small. 

Since YOLOv4 and YOLOv5 are the closest competitors of our proposed approach, we have compared YOLOv4, YOLOv5 and the proposed method further based on OIDv6 dataset. We perform experiment using 80\%-20\% training-testing split on 4038 Billboard class images, where Billboard class contains both advertisement billboard and establishment signboard objects. We did not perform any hyperparameter tuning on this dataset. The scores for YOLOv4, YOLOv5 and the proposed method have been provided in Table \ref{OID}. Regardless of characteristic dissimilarities between billboard and signboard, the end to end training through our proposed pipeline uplifted the test $mAP$ by $9\%$ and $4\%$ compared to YOLOv4 and YOLOv5, respectively.

\begin{table}[h]
  \begin{center}
  \textbf{\caption{ Performance analysis of proposed algorithm with state-of-the-art algorithm based on test set of constructed dataset and existing dataset \label{OID}}}
  \vspace*{1mm}
\resizebox{0.7\textwidth}{!}{%
\begin{tabular}{|c|c|l|}
\hline
Benchmark Dataset                      & Localization Algorithm & \multicolumn{1}{c|}{mAP} \\ \hline
OIDv6 Billboards & YOLOv4                 & 0.77                     \\ \cline{2-3} 
                                       & YOLOv5     & 0.82                     \\ \cline{2-3} 
                                       & Proposed Algorithm     & 0.86                     \\ \hline
SVSO                  & YOLOv4                 & 0.86                     \\ \cline{2-3} 
 & YOLOv5                 & 0.88                     \\ \cline{2-3} 
                                       & Proposed Algorithm     & 0.90                     \\ \hline
\end{tabular}}
\end{center}
\end{table}

\begin{figure}[h]
    \centering
    \includegraphics[width=\linewidth]{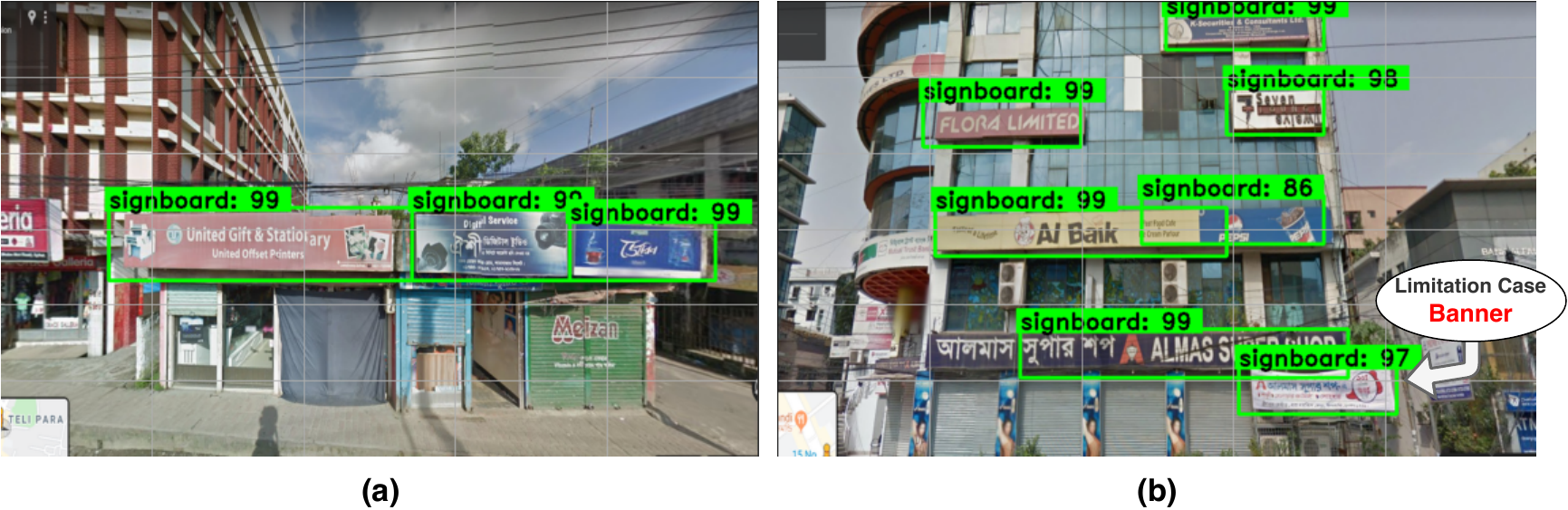}
    \caption{Detection result of our final model on sample SVSO test set images: \textbf{(a)}: signboards detected and localized accurately \textbf{(b)}: a banner has been wrongly classified as signboard.}
    \label{fig:test}
\end{figure}

We have also compared our method with YOLOv4, YOLOv5 on SVSO independent test set (see $4^{th}$, $5^{th}$, and $6^{th}$ row of Table \ref{OID}). The SVSO independent test set contains 200 natural scene images from six diverse places of different countries such as - Bangladesh, Malaysia, Bhutan, Sri Lanka, Myanmar and Thailand. The images have been collected from completely separate places compared to training and validation images. Textual content of the test signboards are written in English, Malay, Burmese, Sinhala, Dzongkha, Bengali and Thai Language. Most of these images have multiple signboards on them. Thus the test set contains 329 signboard samples from 200 image files. The $mAP$ score for our test set is 0.90 Such result proves the generalization ability of our final model in terms of signboard detection and localization of densely populated developing city streets. We show signboard detection result of our proposed model on two sample test images in \textbf{Figure} \ref{fig:test}.

Banners and billboards degrade the signboard detection performance of our proposed model when they show resemblance to signboards with same texture, colour and shapes. \textbf{Figure} \ref{fig:test}b shows such a case. Here, a white banner has been mistakenly detected as signboard. Giving attention to this special case degrades general case performance. So, we leave banner distinction as future research. The proposed training methods are expected to be effective in any Faster R-CNN based localization task. The constructed dataset can be helpful for the scientific community conducting researches in similar area.

\section{Conclusion}
In this research, we have developed an automatic signboard detection and localization system for developing cities. In order to tackle challenges such as multiple signboard detection, heterogeneous background, variation in signboard shapes and sizes, we introduce a robust end to end trainable Faster R-CNN based model integrating new pretraining schemes and hyperparamater selection method. This novel object detection approach achieves 0.90 $mAP$ score on SVSO independent test set and 0.86 $mAP$ score on OIDv6 dataset. This is a significant improvement over state-of-the-art methods such as Faster R-CNN, YOLOv4, and YOLOv5. We have constructed a suitable signboard dataset for this research which we have made available for further research in similar domains. This research is expected to facilitate automatic establishment annotation useful for modern web based location services. Future research may aim at information retrieval task from signboard images, where image resolution is low and signboards contain multiple colour variations. This problem is challenging especially in the context of densely populated developing city streets.

\bibliographystyle{splncs04}
\bibliography{main}

\end{document}